\documentclass[conference]{IEEEtran}
\usepackage{amsmath,amssymb,amsfonts,amsthm}
\usepackage{graphicx}
\usepackage{xcolor}
\usepackage{tikz}
\usetikzlibrary{arrows.meta,shapes.geometric,shadows.blur,positioning,calc}
\usetikzlibrary{arrows.meta,shadows.blur,positioning,fit,calc,decorations.pathmorphing}
\usepackage[hidelinks]{hyperref}

\definecolor{colInput}{HTML}{4E79A7}
\definecolor{colGen}{HTML}{59A14F}
\definecolor{colDecision}{HTML}{F28E2B}
\definecolor{colMitigate}{HTML}{E15759}
\definecolor{colFinal}{HTML}{9C755F}
\definecolor{colYes}{HTML}{2CA02C}
\definecolor{colNo}{HTML}{D62728}
\definecolor{colBg}{HTML}{F7F9FB}

\tikzset{
  font=\small,
  process/.style={
    rectangle, rounded corners=6pt,
    minimum width=3.2cm, minimum height=1.5cm,
    inner sep=6pt, align=center,
    draw=#1!60!black, very thick,
    top color=#1!8, bottom color=#1!30,
    blur shadow={shadow blur steps=5,shadow xshift=1.5pt,shadow yshift=-1.5pt}
  },
  process/.default=colGen,
  decision/.style={
    diamond, aspect=2.2,
    inner sep=1.5pt, align=center,
    draw=colDecision!60!black, very thick,
    top color=colDecision!5, bottom color=colDecision!35,
    text width=2.6cm,
    blur shadow={shadow blur steps=5,shadow xshift=1.5pt,shadow yshift=-1.5pt}
  },
  io/.style={process=colInput},
  gen/.style={process=colGen},
  mit/.style={process=colMitigate},
  final/.style={process=colFinal},
  flow/.style={-Latex, line width=1.2pt, draw=#1},
  flow/.default=black,
  dashedflow/.style={-Latex, line width=1pt, dashed, draw=#1},
  annot/.style={align=left, text width=3.2cm, font=\scriptsize, inner sep=1pt},
  labelbox/.style={rectangle, draw=black!15, fill=black!5, rounded corners=4pt, inner sep=3pt, font=\scriptsize}
}

\tikzstyle{block} = [rectangle, draw, rounded corners, align=center, text width=6em, minimum height=3em]
\tikzstyle{diamondblock} = [diamond, aspect=1.5, draw, align=center, inner sep=1pt, text width=5em]
\tikzstyle{line} = [draw, -latex']

\title{Theoretical Foundations and Mitigation of Hallucination in Large Language Models}
\author{\IEEEauthorblockN{Esmail Gumaan\textsuperscript{*}}
\IEEEauthorblockA{Department of Computer Science, University of Sana'a\\
\href{mailto:23160148@ue.ye.eud}{\texttt{esmailG231601@ue.ye.eud}}}
\thanks{\textsuperscript{*}Corresponding author.}}

\begin{document}
\maketitle

\begin{abstract}
Hallucination in Large Language Models (LLMs) refers to the generation of content that is not faithful to the input or the real-world facts. This paper provides a rigorous treatment of hallucination in LLMs, including formal definitions and theoretical analyses. We distinguish between intrinsic and extrinsic hallucinations, and define a \textit{hallucination risk} for models. We derive bounds on this risk using learning-theoretic frameworks (PAC-Bayes and Rademacher complexity). We then survey detection strategies for hallucinations, such as token-level uncertainty estimation, confidence calibration, and attention alignment checks. On the mitigation side, we discuss approaches including retrieval-augmented generation, hallucination-aware fine-tuning, logit calibration, and the incorporation of fact-verification modules. We propose a unified detection and mitigation workflow, illustrated with a diagram, to integrate these strategies. Finally, we outline evaluation protocols for hallucination, recommending datasets, metrics, and experimental setups to quantify and reduce hallucinations. Our work lays a theoretical foundation and practical guidelines for addressing the crucial challenge of hallucination in LLMs.
\end{abstract}

\section{Introduction}
Large Language Models (LLMs) such as GPT-3 have demonstrated remarkable capabilities in natural language generation, achieving fluent and contextually relevant outputs in tasks from summarization to dialogue \cite{Brown2020, Devlin2019, Vaswani2017}. However, a critical challenge that has emerged with these models is the tendency to \textit{hallucinate}—produce plausible-sounding content that is factually incorrect or not supported by the input or reality \cite{Ji2023, Maynez2020}. Hallucinations can manifest in various forms, from minor factual inaccuracies to entire fabricated statements, undermining the reliability of LLMs in high-stakes applications (e.g., medical or legal domains) \cite{Farquhar2024}. The prevalence of hallucinations has been observed across multiple domains: for instance, early neural dialogue systems sometimes generated inconsistent or untrue responses \cite{Vinyals2015}, neural machine translation systems occasionally produced unrelated outputs especially for out-of-distribution inputs \cite{Koehn2017, Raunak2021}, and abstractive summarization models often include details not present in the source text \cite{Maynez2020}. Researchers have broadly categorized hallucinations in text generation into two types: \textit{intrinsic} and \textit{extrinsic} \cite{Maynez2020, Ji2023}. Intrinsic hallucinations occur when the generated output contradicts or distorts the given source input, while extrinsic hallucinations introduce new information that cannot be verified against the source (often introducing facts that are entirely fabricated or irrelevant) \cite{Maynez2020}. Both types are problematic, with intrinsic hallucinations violating input faithfulness and extrinsic hallucinations potentially spreading misinformation if taken as factual \cite{Ji2023}. Recent surveys underscore that hallucination is a pervasive issue in current LLMs \cite{Ji2023, Zhang2023}, and significant research efforts are focused on understanding and mitigating this phenomenon.

In this paper, we undertake a comprehensive exploration of the theoretical foundations of hallucination in LLMs and the strategies to detect and mitigate it. We begin by providing formal definitions of hallucinations, distinguishing between intrinsic and extrinsic cases in mathematical terms. Building on these definitions, we introduce the notion of \textit{hallucination risk} for a language model and derive theoretical bounds on this risk. In particular, we leverage tools from statistical learning theory, including PAC-Bayesian analysis and Rademacher complexity, to bound the probability of hallucination under certain assumptions. This theoretical perspective clarifies the limits of learning and highlights why completely eliminating hallucinations may be inherently difficult in general settings (echoing recent results that suggest hallucinations are fundamentally inevitable in sufficiently complex models \cite{Xu2024}).

We then shift to practical aspects: first, we survey methods for detecting hallucinations in LLM outputs. These include token-level uncertainty estimation techniques (e.g., using probability distributions or entropy to flag low-confidence predictions \cite{Gal2016, Kadavath2022}), confidence calibration methods to adjust a model's reported confidence to better reflect factual accuracy \cite{Guo2017, Jiang2021}, and attention-based checks that verify whether generated content is properly grounded in source input via the model's attention patterns (particularly for tasks like summarization or translation) \cite{See2017}. Next, we discuss mitigation strategies to reduce hallucinations during generation. Among these, retrieval-augmented generation (RAG) has emerged as a powerful approach, wherein the model consults an external knowledge source to ground its responses in factual data \cite{Lewis2020, Guu2020, Shuster2021}. We also cover hallucination-aware fine-tuning and reinforcement learning strategies that train models to avoid unsupported content \cite{Ouyang2022, Nakano2022}, as well as techniques like logit calibration (adjusting the model’s output probabilities or decoding strategy to prevent overconfident leaps) \cite{Desai2020, Holtzman2020}. Another important direction is augmenting LLMs with fact-verification heads or modules that cross-check generated statements against reference knowledge bases or learned factual representations \cite{Kryscinski2020, Manakul2023}. By integrating a verification step, the model can potentially catch and correct hallucinations before presenting the final output.

We integrate these insights by proposing a unified workflow for hallucination detection and mitigation. The workflow involves an LLM generating a preliminary answer, a detection module assessing the answer’s fidelity (through uncertainty signals and content verification), and a mitigation step (such as retrieving relevant facts or adjusting the answer) if a potential hallucination is detected. We present a diagram illustrating this pipeline and discuss how each component interacts.

Finally, we address the evaluation of hallucinations and their mitigation. We outline recommended datasets and benchmarks for assessing factuality and faithfulness, such as TruthfulQA for open-domain truthfulness \cite{Lin2021}, factual consistency benchmarks in summarization (e.g., datasets with human-annotated hallucinations \cite{Maynez2020, Pagnoni2021}), and domain-specific tests (like medical question answering where factual accuracy is critical). We also review common metrics for quantifying hallucinations, ranging from simple precision/recall of factual statements to more sophisticated automatic metrics like FactCC \cite{Kryscinski2020}, QAGS \cite{Wang2020}, and TRUE \cite{Honovich2022}. We emphasize the importance of human evaluation as the gold standard for detecting subtle hallucinations and recommend experimental protocols (such as A/B testing of models with and without mitigation modules, and measuring improvements in factual accuracy and calibration).

The remainder of this paper is organized as follows. Section~\ref{sec:definition} formalizes the definition of hallucination in language generation, including intrinsic and extrinsic forms, and introduces the hallucination risk framework. Section~\ref{sec:theory} develops theoretical bounds on hallucination risk using PAC-Bayes and complexity measures. Section~\ref{sec:detection} discusses approaches for detecting hallucinations in LLM outputs. Section~\ref{sec:mitigation} presents various mitigation techniques to reduce hallucinations. In Section~\ref{sec:workflow}, we propose a combined detection-mitigation workflow and provide a diagrammatic representation. Section~\ref{sec:evaluation} gives recommendations for evaluation datasets, metrics, and experimental procedures to study hallucinations. Finally, Section~\ref{sec:conclusion} concludes with reflections on future research directions for making LLMs more truthful and reliable.

\section{Hallucination Definitions and Formalization}
\label{sec:definition}
In this section, we provide a formal definition of hallucination in the context of language models. We distinguish between \textit{intrinsic} hallucinations (inconsistencies with respect to a given source input) and \textit{extrinsic} hallucinations (content unsupported by any provided input or known reference), following the taxonomy introduced in the summarization and translation literature \cite{Maynez2020, Ji2023}. 

\subsection{Intrinsic vs. Extrinsic Hallucinations}
Consider a conditional language generation setting with input $x$ (e.g., a source document or prompt) and output text $y$ produced by the model. Let $I(x)$ denote the set of factual assertions present in the input $x$. Likewise, let $O(y)$ denote the set of assertions made in the output $y$. We say that the output $y$ is \textbf{intrinsically hallucinated} (with respect to $x$) if there exists at least one proposition in $O(y)$ that \emph{directly contradicts} information in $I(x)$. In other words, $y$ asserts something that is negated or refuted by the source content $x$ (thus violating consistency) \cite{Maynez2020}. Formally:
\[
\exists p \in O(y) \text{ such that } p \text{ is logically incompatible with } I(x).
\]
Intrinsic hallucinations are often easier to detect because the contradiction with the source can sometimes be identified by entailment-checking or overlap with source facts. For example, if $x$ is a document stating "The Eiffel Tower is located in Paris" and the summary $y$ states "The Eiffel Tower is located in Rome," $y$ contains an intrinsic hallucination (a directly contradictory claim).

In contrast, $y$ is said to be \textbf{extrinsically hallucinated} if $y$ includes information that is not present in $I(x)$ and cannot be verified by any accessible knowledge source, despite not necessarily contradicting $x$ \cite{Maynez2020, Ji2023}. Extrinsic hallucinations introduce \emph{new} assertions that go beyond the input. Formally:
\[
\exists q \in O(y) \text{ such that } q \not\vdash I(x) \text{ (not entailed by $I(x)$)},
\] 
and typically $q$ corresponds to some factual claim for which $x$ provides no evidence. For example, if $x$ is an article about Paris and $y$ (a summary) adds a sentence "Paris is home to the largest rainforest in Europe," this is extrinsic hallucination: the added detail is not in the source and is in fact a fabricated or unrelated claim (and cannot be verified as true from the given input). Extrinsic hallucinations often require external knowledge or fact-checking to detect, since the model might introduce a plausible-sounding but incorrect fact that the input never mentioned.

It is worth noting that whether a piece of generated content counts as a hallucination can depend on the task context and the expected scope of the output. In strictly input-bound tasks (like translation or faithful summarization), \emph{any} content not grounded in the input is undesired (and thus extrinsically hallucinated). In open-ended creative generation or dialogue, extrinsic additions might be acceptable or even required (for engagement), as long as they remain consistent with general world knowledge and do not introduce false facts. In this work, we focus on hallucinations in contexts where factuality and faithfulness are expected, such as summarization, question-answering, and knowledge-grounded dialogue.

\subsection{Hallucination Risk}
We introduce the notion of \textit{hallucination risk} to quantify how prone a language model is to hallucinate. Intuitively, hallucination risk refers to the probability that the model's output will contain a hallucination (of either type) under the distribution of inputs of interest. 

Let $\mathcal{X}$ be a distribution over inputs (and possibly paired with ground-truth outputs or an underlying truth source). We assume there is a (possibly unknown) ground-truth function or oracle $f^*(x)$ that provides a fully truthful and contextually appropriate output for input $x$ (for instance, the correct answer in a QA task or a perfectly faithful summary of a document). We can then define a \textbf{hallucination indicator} for a model $M$ on input $x$:
\[
H(M, x) = 
\begin{cases}
1, & \text{if the output $y = M(x)$ hallucination}, \\
0, & \text{if $M(x)$ is completely faithful to $f^*(x)$ (no hallucination).}
\end{cases}
\]
Using this indicator, the \textbf{hallucination risk} $R_{\mathrm{hall}}(M)$ is the expected value:
\[
R_{\mathrm{hall}}(M) = \mathbb{E}_{x \sim \mathcal{X}}\left[ H(M, x) \right],
\] 
which is simply the probability that $M$ produces a hallucination on a random input from $\mathcal{X}$. In practice, since hallucination is a binary condition per output, this risk can also be interpreted as the hallucination rate.

We can similarly define an \textit{intrinsic hallucination risk} and \textit{extrinsic hallucination risk} if we wish to separate the two types:
\[
R_{\mathrm{int}}(M) = \Pr_{x}[ M(x) \text{ has an intrinsic hallucination}],
\] 
\[
R_{\mathrm{ext}}(M) = \Pr_{x}[ M(x) \text{ has an extrinsic hallucination}].
\]
These are useful if the application cares differently about each type (for example, in summarization, intrinsic hallucinations might indicate serious errors against source fidelity, whereas extrinsic hallucinations might be less damaging if they are minor details, though both are generally undesirable).

Hallucination risk depends on both the model and the input distribution. A model might hallucinate rarely on inputs drawn from in-domain data (where it was trained or where it has strong knowledge) but hallucinate more on out-of-distribution or open-domain inputs where its knowledge is uncertain. This aligns with the intuition that hallucinations often occur when the model is faced with queries that exceed its knowledge or stray from the support of its training data.

\section{Theoretical Analysis of Hallucination Risk}
\label{sec:theory}
Given the formalization above, we now consider theoretical bounds on hallucination risk. We can treat hallucination detection as a binary classification problem on model outputs (hallucinated vs. faithful). If we have a training dataset or some feedback that labels when outputs are hallucinations, one can in principle train a model or adjust the original model to minimize this risk. The challenge is that the space of all possible outputs is enormous, and directly supervising every case of hallucination is infeasible. Nonetheless, we can leverage generalization bounds from learning theory to reason about the hallucination behavior of models.

\subsection{Generalization Perspective}
From a learning perspective, a language model $M$ can be seen as attempting to approximate the ground truth function $f^*$ (which produces fully factual outputs) based on finite training data. Hallucinations are then instances of generalization error: cases where $M(x)$ deviates from $f^*(x)$ in a way that introduces false or unsupported content. In fact, one might say that the ultimate goal of an ideal training procedure (for tasks requiring factuality) is to minimize $R_{\mathrm{hall}}(M)$.

If we had a way to automatically determine whether a given output is hallucinatory (for example, via human labeling or a fact-checking oracle), we could measure an empirical hallucination rate on a sample of $n$ inputs:
\[
\hat{R}_{\mathrm{hall}}(M) = \frac{1}{n} \sum_{i=1}^n H(M, x_i),
\] 
where $x_1, \dots, x_n$ are sample inputs (and we assume we can identify if $M(x_i)$ hallucinated). This is analogous to an empirical risk (error rate) in binary classification where "error" corresponds to hallucinating.

Standard generalization bounds would then relate $R_{\mathrm{hall}}(M)$ to $\hat{R}_{\mathrm{hall}}(M)$. For example, if $M$ comes from a hypothesis class of bounded complexity (e.g., limited capacity or effectively controlled by regularization), we can invoke a uniform convergence bound. Using a VC-dimension or Rademacher complexity argument \cite{Bartlett2002, Vapnik1998}, one would state that with high probability (over the choice of the $n$ inputs):
\[
R_{\mathrm{hall}}(M) \le \hat{R}_{\mathrm{hall}}(M) + \mathcal{O}\!\Big(\sqrt{\frac{\mathcal{C}}{n}}\Big),
\] 
where $\mathcal{C}$ is a measure of model complexity (e.g., VC-dimension or a bound on Rademacher complexity of the associated hypothesis class), and the big-O hides constant factors and $\ln(1/\delta)$ terms for a given confidence $1-\delta$. In simpler terms, if a model exhibits a low hallucination frequency on a representative training set and the model class is not too complex, we expect it to have a low hallucination probability on new inputs as well.

However, modern LLMs are extremely high-capacity (often overparameterized) models, which makes classical complexity measures very large. In practice, they can have near-zero training error but still hallucinate on new inputs, meaning the challenge is often one of distribution shift or incomplete knowledge rather than traditional generalization in the statistical learning sense.

\subsection{PAC-Bayesian Bound on Hallucination Risk}
Another way to derive a bound is through the PAC-Bayes framework, which is well-suited for reasoning about the generalization of complex models by introducing a prior and considering a distribution over models \cite{McAllester1999}. We can derive a PAC-Bayes bound on the hallucination risk as follows. Assume a prior distribution $P$ over models (e.g., before observing any data, some distribution reflecting our initial belief about model parameters) and let $Q$ be a posterior distribution (concentrated around the trained model). For any $\delta > 0$, with probability at least $1-\delta$ over the random draw of the training data, the following bound holds for all distributions $Q$ (over models in our hypothesis space):
\begin{equation}
\mathbb{E}_{M \sim Q}[R_{\mathrm{hall}}(M)] \le \mathbb{E}_{M \sim Q}[\hat{R}_{\mathrm{hall}}(M)] + \sqrt{\frac{\mathrm{KL}(Q||P) + \ln\frac{1}{\delta}}{2n}}, \label{eq:pacbayes}
\end{equation}
where $\mathrm{KL}(Q||P)$ is the Kullback-Leibler divergence between $Q$ and $P$. This is a direct application of the PAC-Bayesian generalization bound for 0-1 loss (which hallucination indicator essentially is) \cite{McAllester1999}. If we take $Q$ to be a point mass at our learned model $M$ (i.e., we are considering the deterministic model we have after training), the bound simplifies to an upper bound on $R_{\mathrm{hall}}(M)$ in terms of the empirical hallucination rate of $M$ plus a complexity penalty that scales with the description length of $M$ relative to the prior and with $1/\sqrt{n}$.

The utility of the bound in (\ref{eq:pacbayes}) is mostly conceptual for our purposes: it tells us that if a model has low hallucination rate on the training data (perhaps through fine-tuning on high-quality, factual responses) and if the model is not overly complex relative to our prior beliefs, then we can guarantee a low hallucination risk on new data, with high probability. Of course, in practice, defining a sensible prior $P$ and computing the KL term can be challenging for large neural networks. Nonetheless, PAC-Bayesian analysis has been used to explain generalization even in overparameterized models by choosing informative priors (e.g., centered at an earlier state of the model before fine-tuning).

\subsection{On the Impossibility of Eliminating Hallucination Completely}
A recent theoretical result suggests that for sufficiently powerful models, some degree of hallucination may be fundamentally unavoidable \cite{Xu2024}. In a formal setting, one can prove that no computable model can perfectly reproduce another arbitrary computable ground-truth function $f^*$ in all cases (this is related to results in computational learning theory and the limitations of generalization). Informally, if an LLM is used as a general problem solver across an open-ended space of queries, there will always be some inputs for which the model fails to produce the correct output (unless the model is as powerful as the oracle $f^*$ itself, which in realistic terms it is not). These failures manifest as hallucinations when the model still produces an answer, but that answer is not the correct or truthful one. In other words, hallucination in extremely general settings can be seen as a consequence of the fact that LLMs cannot know everything or perfectly generalize to every possible query \cite{Xu2024}. This aligns with intuition: an LLM that has not been exposed to a particular rare fact during training might “guess” and thereby hallucinate when asked about it.

The theoretical takeaway is that while we can reduce the probability of hallucination (and aim to make it very low, especially in critical applications), completely eliminating hallucinations for all possible inputs is likely infeasible. Thus, detection and mitigation strategies (discussed next) are crucial complements to training better models.

\section{Detection of Hallucinations}
\label{sec:detection}
Before we can mitigate or prevent hallucinations, we must detect when they occur or are likely to occur. Detection can happen \textit{post hoc} (after a model generates an output, identify if it contains a hallucination) or \textit{online} (during generation, identify tokens or sequences that are potentially hallucinated in real-time). We explore several approaches to hallucination detection in LLM outputs, focusing on:
\begin{itemize}
    \item \textbf{Uncertainty and token-level cues} – methods that use the model's own predicted probabilities or variations to gauge confidence.
    \item \textbf{Confidence calibration and self-evaluation} – methods where the model or an auxiliary model assesses the likelihood of its output being correct.
    \item \textbf{Attention and attribution-based checks} – methods that inspect whether the model’s output content is properly grounded in the input via alignment techniques.
\end{itemize}

\subsection{Token-Level Uncertainty Estimation}
One indicative signal of a potential hallucination is the model's \emph{uncertainty} in generating certain tokens or facts. Intuitively, if the model is not confident (according to its own probability distribution) about a particular piece of generated information, that piece may be a hallucination. However, raw probabilities from a language model are not always well-calibrated (a model might assign high probability to a guess due to learned patterns, not because it is certain of factual correctness) \cite{Kadavath2022}. 

Approaches to quantify uncertainty at the token or sequence level include:
\begin{itemize}
    \item \textbf{Entropy or variance of predictions:} The entropy of the model's next-token distribution is a measure of uncertainty. High entropy means the model is unsure which token to produce next. If a model produces a token (or sequence) while the entropy is high, it might be "guessing," which can correlate with hallucination. Similarly, one can sample multiple continuations from the model (via Monte Carlo dropout or an ensemble of model snapshots \cite{Gal2016}) and measure variance among outcomes. A high variance in answers to the same prompt often signals low confidence in any single answer \cite{Farquhar2024}.
    \item \textbf{Consistency under perturbations:} A method known as \textit{self-consistency} involves posing the same question or prompt to the model multiple times (or with slight rephrasings or different sampling seeds) and seeing if the model’s answer remains consistent. If the model's answers fluctuate significantly (especially on factual questions) this can indicate it does not actually "know" the answer and might be confabulating \cite{Farquhar2024}. For example, if asked "What is the target of Sotorasib" and the model sometimes answers one protein and other times another despite identical instructions, it suggests hallucination risk. 
    \item \textbf{Surprise relative to training data:} Another angle is to measure how likely a generated statement is under a reference distribution of truthful statements (for example, using a smaller fact-grounded model or n-gram statistics). If a sentence in $y$ contains a very rare or unprecedented combination of tokens that was not seen in truthful contexts, it could be flagged.
\end{itemize}

In practice, Farquhar \textit{et al.} (2024) propose computing uncertainty at the level of semantic content by clustering paraphrases of the output and measuring entropy in the semantic space, which they found effective in detecting what they call "confabulations" (arbitrary, incorrect answers) \cite{Farquhar2024}. This method aims to overcome the limitation that one idea can be expressed in many lexical ways – instead of surface-level entropy, they consider if the model is uncertain about the meaning it wants to convey.

Token-level uncertainty methods are appealing because they do not require external data; they use the model's own behavior as a signal. However, not all hallucinations come with obvious uncertainty — sometimes a model will assert a wrong fact with high confidence (low entropy), which is a worst-case scenario. This is where calibration and external checks become vital.

\subsection{Confidence Calibration and Self-Evaluation}
Confidence calibration refers to the process of adjusting or interpreting the model's output probabilities so that they reflect the true likelihood of correctness \cite{Guo2017}. An uncalibrated model might be overconfident in false outputs or underconfident in true outputs. Calibrating a language model's confidence could involve:
\begin{itemize}
    \item \textbf{Temperature scaling:} Applying a softmax temperature or isotonic regression on predicted probabilities to better align them with empirical correctness likelihoods \cite{Guo2017}. For example, one might fine-tune a model on a set of QA examples with known true/false answers to calibrate the relationship between the model’s probability distribution and whether the answer was correct.
    \item \textbf{Ask the model to rate its confidence:} Some works have shown that LLMs can produce a qualitative self-assessment when prompted (e.g., "How sure are you about the above answer"). While not entirely reliable, in certain settings large models can indicate uncertainty (like by saying "I'm not entirely sure") which correlates with actual error \cite{Kadavath2022}. There are also methods where the model is trained or prompted to output a probability or confidence score along with the answer \cite{Jiang2021}.
    \item \textbf{Calibrating through few-shot examples:} Providing a few examples of answers with confidence levels (or probabilities of being correct) in a prompt can sometimes calibrate the model in a zero-shot or few-shot manner. This instructs the model on how it should distribute probability mass when unsure.
    \item \textbf{External calibration models:} We can train a separate model (or use a smaller verifier network) that takes as input the LLM's output (and possibly the question or context) and predicts a probability that the output is correct. This is akin to a regression or classification (true vs false) on the content of the answer. If well-trained, such a model can effectively flag likely hallucinations by outputting a low score for unfaithful answers \cite{Kryscinski2020, Honovich2022}.
\end{itemize}

Confidence-based detectors often yield a scalar "factuality score" or likelihood of correctness for a given output. For example, a verification model might be built using a dataset of known factual vs hallucinated outputs and then applied to new outputs (similar to how Fact was a BERT-based classifier trained to detect factual consistency of summaries \cite{Kryscinski2020}). A well-calibrated system would ideally either abstain (choose not to answer) or indicate uncertainty when it is likely to hallucinate, thereby preventing misinformation. Techniques like selective prediction \cite{Kamath2020} implement this: the system only outputs an answer when it is confident it is correct, otherwise it says "I don't know" or defers.

\subsection{Attention Alignment and Source Attribution}
For tasks where the model is provided with a source (e.g., document, knowledge base, or context), we can exploit the model's internal attention or alignment mechanisms to detect hallucinations. The intuition is that for a factual statement in the output, there should be some part of the input or retrieved context that the model attended to or based that statement on. If the model outputs a sentence that has no corresponding source span and the model's attention distribution during generation of that sentence did not focus on any relevant input tokens, this could be a sign of hallucination (specifically extrinsic hallucination).

Several methods in this vein:
\begin{itemize}
    \item \textbf{Attention-based provenance:} For each token or phrase the model generates, one can look at the encoder-decoder attention weights (in a seq2seq model like a Transformer) to see which input tokens influenced it. If a noun phrase or a factual assertion in the output has uniformly low attention weights across all input tokens (or attends to irrelevant parts of the input), it suggests the model is "hallucinating" that content without grounding \cite{See2017}. For example, in translation, if the model outputs a phrase that was never in the source and attention doesn't align it to any source phrase, it's likely a hallucination.
    \item \textbf{Gradient or attribution methods:} Beyond raw attention, one can use input attribution techniques (like integrated gradients or attention flow) to identify which parts of the input most contributed to the generation of a specific output segment. Hallucinated content would show weak attribution to input features, whereas faithful content should trace back to some input evidence.
    \item \textbf{Verification with retrieval:} This approach is related to mitigation but can be used purely for detection: given a model output sentence, issue it (or its claim) as a query to a search engine or knowledge base. If no supporting document or evidence can be found, flag it as potential hallucination. This is essentially how a human fact-checker might operate. While not an internal model method, it is a practical detection strategy. Research prototypes have used web search to detect likely factual errors in model outputs by seeing if the facts appear in credible sources.
\end{itemize}

For summarization tasks, metrics like entity overlap or content coverage have been used: e.g., count how many named entities in the summary appear in the source. A summary that introduces unseen entities or numbers is likely hallucinating extrinsic details \cite{Maynez2020}. Similarly, in knowledge-grounded dialogue, if the conversation is supposed to stay grounded in provided knowledge snippets and the model response introduces a new factoid not in the snippets, a detection system can catch that by string matching or semantic matching against the knowledge source \cite{Shuster2021}. 

Attention alignment checks have their limitations: high attention weight doesn't guarantee correctness (the model could attend to the right source but still generate an incorrect interpretation of it), and low attention weight might sometimes be deceiving (since attention can be diffused). Nonetheless, combined with other signals, attention patterns are a useful indicator.

\section{Mitigation Strategies for Hallucination}
\label{sec:mitigation}
We now discuss strategies to mitigate (reduce or prevent) hallucinations in LLMs. These methods can be applied during the model training phase, at decoding time, or as post-processing steps, and many are complementary (they can be combined for better results). We focus on four broad categories that have shown promise:

\subsection{Retrieval-Augmented Generation (RAG)}
One of the most effective ways to ground a language model in factual content is to provide it with relevant reference information at generation time \cite{Lewis2020}. Retrieval-Augmented Generation (RAG) frameworks augment the model with a retrieval mechanism: given an input (e.g., a question), the system first retrieves documents or knowledge from a large external database or the web, and then the language model conditions its generation on both the input and the retrieved evidence. Because the model has access to explicit knowledge, it is less likely to fill gaps with hallucinated facts from its parameters; instead, it can quote or fuse information from the retrieved text.

Key aspects of RAG include:
\begin{itemize}
    \item \textbf{Training with retrieval:} Some models like REALM \cite{Guu2020} and RETRO incorporate retrieval directly into training. They learn to use a differentiable retrieval component such that at inference time, they continue to retrieve relevant text for each query. These models have been shown to produce more accurate, factual statements since they can look up facts rather than rely purely on memorized knowledge.
    \item \textbf{Open-domain QA and knowledge-grounded dialogue:} RAG has been successfully applied in open-domain question answering, where models like DrQA and RAG \cite{Lewis2020} retrieve Wikipedia articles to answer questions, dramatically reducing the chance of unsupported answers. Similarly, dialog systems (like BlenderBot 2.0) retrieve knowledge to ground their responses, thereby mitigating hallucinations in conversation \cite{Shuster2021}.
    \item \textbf{Plug-and-play retrieval modules:} Even if the base LLM is not trained with retrieval, one can implement a pipeline: first retrieve top-$k$ relevant documents (using an IR system or a dense retriever), then prepend or concatenate those documents to the model's input context, and finally generate the output. This provides the model with the opportunity to copy or use actual facts from the evidence, rather than guessing. Many recent LLM applications (e.g., Bing's chat or other QA systems) use this approach to improve factual accuracy.
    \item \textbf{Challenges:} RAG is not a panacea; if the retrieval fails (e.g., no relevant document is found) or if the model misinterprets the retrieved text, hallucinations can still occur. There is also the risk of the model citing a retrieved fact incorrectly. However, the overall empirical finding is that grounding generation in retrieved data significantly reduces hallucination rates in tasks like QA, as long as a correct reference can be found \cite{Lewis2020}.
\end{itemize}

By anchoring the generation to external knowledge, RAG effectively shifts the problem from "the model must know everything" to "the model must know how to find and use information," which is easier to achieve reliably. The model becomes an assembler of facts from its sources rather than a sole source of facts.

\subsection{Hallucination-Aware Fine-Tuning and Instruction Tuning}
Another approach is to train the model explicitly to avoid hallucinations. This can be done through fine-tuning on datasets that emphasize factual correctness, or via reinforcement learning with feedback on factuality.

Some methods include:
\begin{itemize}
    \item \textbf{Supervised fine-tuning on truthful data:} If we have high-quality datasets of question-answer pairs, summaries, or dialogues where the output is guaranteed to be factual (and we have negatives that are hallucinated), we can fine-tune the LLM on this data to encourage factual generation. For example, models can be fine-tuned to produce "faithful summaries" by using a dataset of summaries that was cleaned of hallucinations or by adding a loss term that penalizes including content not present in the source.
    \item \textbf{Reinforcement Learning from Human Feedback (RLHF):} RLHF has been used to align LLMs with human preferences, which include truthfulness \cite{Ouyang2022}. In RLHF, the model generates outputs, and a reward model (often trained on human preference data) gives higher scores to outputs that are correct and lower to those that are hallucinated or incorrect. By optimizing for this reward via policy gradients or proximal policy optimization (PPO), the model learns to avoid outputs that humans would label as hallucinations or unhelpful. InstructGPT and related models have used this to reduce the incidence of blatant false statements \cite{Ouyang2022}.
    \item \textbf{Penalizing unsupported content:} One can introduce a training signal that explicitly checks whether each statement in the output is supported by the input (for tasks where input is present). If not, a penalty is applied. This can be done by integrating a differentiable verification mechanism or by data augmentation (provide negative examples of unsupported statements and train the model to output a special token or refrain in those cases).
    \item \textbf{Encouraging refusals for unknowns:} Instruction tuning can teach the model to respond with uncertainty when it doesn't know an answer. For example, including prompts in the fine-tuning data like "Q: [difficult question] A: I'm sorry, I don't have enough information to answer that." helps the model learn that saying "I don't know" is acceptable and preferable to hallucinating \cite{Lin2021}. This strategy directly combats extrinsic hallucinations by essentially opting out of answering when likely to hallucinate.
\end{itemize}

Hallucination-aware training leverages the training process to instill caution and fact-awareness in the model. However, it requires relevant training data or feedback. Human annotation of hallucinations can be expensive, so some works use semi-supervised approaches: e.g., generate candidate outputs and automatically label them as hallucinated or not using a heuristic or another model, then fine-tune on that.

It is also noteworthy that focusing too much on factual correctness can sometimes degrade the model's creativity or ability to generalize (a phenomenon sometimes called the "alignment tax" where making models safer or more factual might reduce some capability \cite{Ji2023}). Thus, fine-tuning must strike a balance, and often it's combined with other methods rather than being the sole solution.

\subsection{Logit Calibration and Decoding Strategies}
Even without additional training or external knowledge, we can often reduce hallucinations by carefully controlling the generation process. Hallucinations are sometimes linked to the model "overshooting" with a highly likely token that leads down a wrong path (especially in greedy or beam search decoding), or conversely, sampling too freely such that random errors slip in. Techniques to calibrate or constrain the model’s logits (the raw probabilities for next tokens) and to adjust decoding can help:
\begin{itemize}
    \item \textbf{Lowering the temperature / Nucleus sampling:} By using a lower temperature in softmax sampling, we make the model’s output distribution more peaky, effectively making it more deterministic and less likely to produce low-probability (potentially nonsensical) tokens. Nucleus (top-$p$) sampling \cite{Holtzman2020} limits the sampling to a subset of tokens that cover a cumulative probability $p$ (often $p=0.9$). This avoids tail tokens that the model assigned small probability, which could be wild off-track continuations. These methods generally improve coherence and relevance, which indirectly reduces hallucinations. However, too low a temperature can also cause repetition or sticking to safe phrases.
    \item \textbf{Ban or penalize unsupported tokens:} If we have some idea of what tokens or phrases are likely to be hallucinated (for example, the model might consistently make up references or URLs in a certain format), we can apply a logit penalty or mask to prevent those from being generated unless certain conditions are met. Some production AI systems maintain blacklists or use detection during generation to stop output if a hallucination pattern is detected (though this is usually more for toxic content, it can be adapted to factual errors).
    \item \textbf{Constrained decoding with knowledge:} Another advanced idea is to integrate a verification step into decoding. For instance, after each sentence generated, an auxiliary checker (like a fact-check model or a retrieval query) could validate the statement. If the checker indicates a likely error, the model can be guided to revise or drop that part. This requires careful orchestration but has been explored in methods allowing an LLM to call external tools mid-generation.
    \item \textbf{Logit adjustment for calibration:} If the model is known to be overconfident (its probability distribution has too low entropy compared to actual uncertainty), one can flatten the distribution (increase entropy) in a targeted way. Conversely, if a model tends to guess when it shouldn't, one might detect such situations (e.g., the query is obscure) and dynamically lower the probability of tokens that would assert facts (like names, dates) to encourage the model to say it doesn't know or to hedge.
\end{itemize}

Holtzman \textit{et al.} \cite{Holtzman2020} showed that typical max-likelihood decoding (greedy or beam search) often yields degenerate repetitive text because it over-commits to high-probability words, whereas sampling can produce more natural text. But pure sampling can also produce off-topic or incorrect info. Nucleus sampling was a compromise ensuring both coherence and diversity. In the context of factuality, one might similarly tune decoding parameters to favor safer, on-distribution completions. A very aggressive strategy is to use beam search but with a re-ranking at the end by a factuality model, selecting the highest plausible answer that is factually consistent.

\subsection{Fact-Verification Modules and Auxiliary Heads}
Finally, one can extend the model architecture by adding components dedicated to factual verification. This can take several forms:
\begin{itemize}
    \item \textbf{Classifier head on the decoder:} For instance, during generation, after each token or sentence, an auxiliary classifier (attached to the model's hidden state) could predict whether the sequence so far is factual and consistent with the source. If it predicts a high probability of hallucination, the model could adjust or a constraint could be applied to steer it back. Training such an internal classifier might involve multi-task learning: the model not only learns to generate the next token but also to judge the truth of what has been generated so far.
    \item \textbf{Two-pass generation (draft and verify):} In this approach, the model first generates a draft answer. Then, a second pass (either by the same model or a specialized verification model) checks each statement in the draft. The model can then be prompted to correct any statements that the verifier flagged as likely incorrect. This chain-of-verification idea is explored by Dhuliawala \textit{et al.} (2023) \cite{Dhuliawala2023}, where the model generates and then formulates verification questions about its own output, answering them with an external tool or its own internal knowledge to validate the content. This effectively adds a self-check mechanism.
    \item \textbf{Tool use and fact-checkers:} LLMs can be augmented to use external tools like search engines, calculators, or knowledge bases mid-generation (e.g., Toolformer techniques). For factual questions, the model can decide to issue a query and then incorporate the result, rather than directly answering from its parametric memory. If integrated properly, this ensures that specific facts are fetched from a reliable source. The model’s architecture might include a decision head that triggers a tool usage when confidence is low.
    \item \textbf{Knowledge incorporation in training:} Another angle is to embed a knowledge graph or database into the model's representations, or to pre-train/fine-tune the model with objectives that tie it to factual knowledge (such as link prediction tasks or masked language modeling on factual texts). A model that has an internal knowledge graph can effectively have a built-in verifier that checks consistency against that graph. Some recent studies incorporate knowledge graph embeddings into the transformer hidden states to bias generation toward known facts \cite{Zhao2023}.
\end{itemize}

The idea of a fact-verification module is analogous to having an editor or fact-checker watch over the writer (the LLM) in real-time. This can catch errors that the writer might make inadvertently. For example, if the model says "In 1975, the population of X was Y," a verification head might internally cross-check that with its training data or an external table and realize it's hallucinated, then either adjust that part or not output it.

An important consideration is that any verification system is only as good as the knowledge it has. If either the model or the verifier lack certain knowledge, they might not catch a hallucination. Thus, combining retrieval (to supply knowledge) with verification is often most effective.

\section{Proposed Detection and Mitigation Workflow}
\label{sec:workflow}
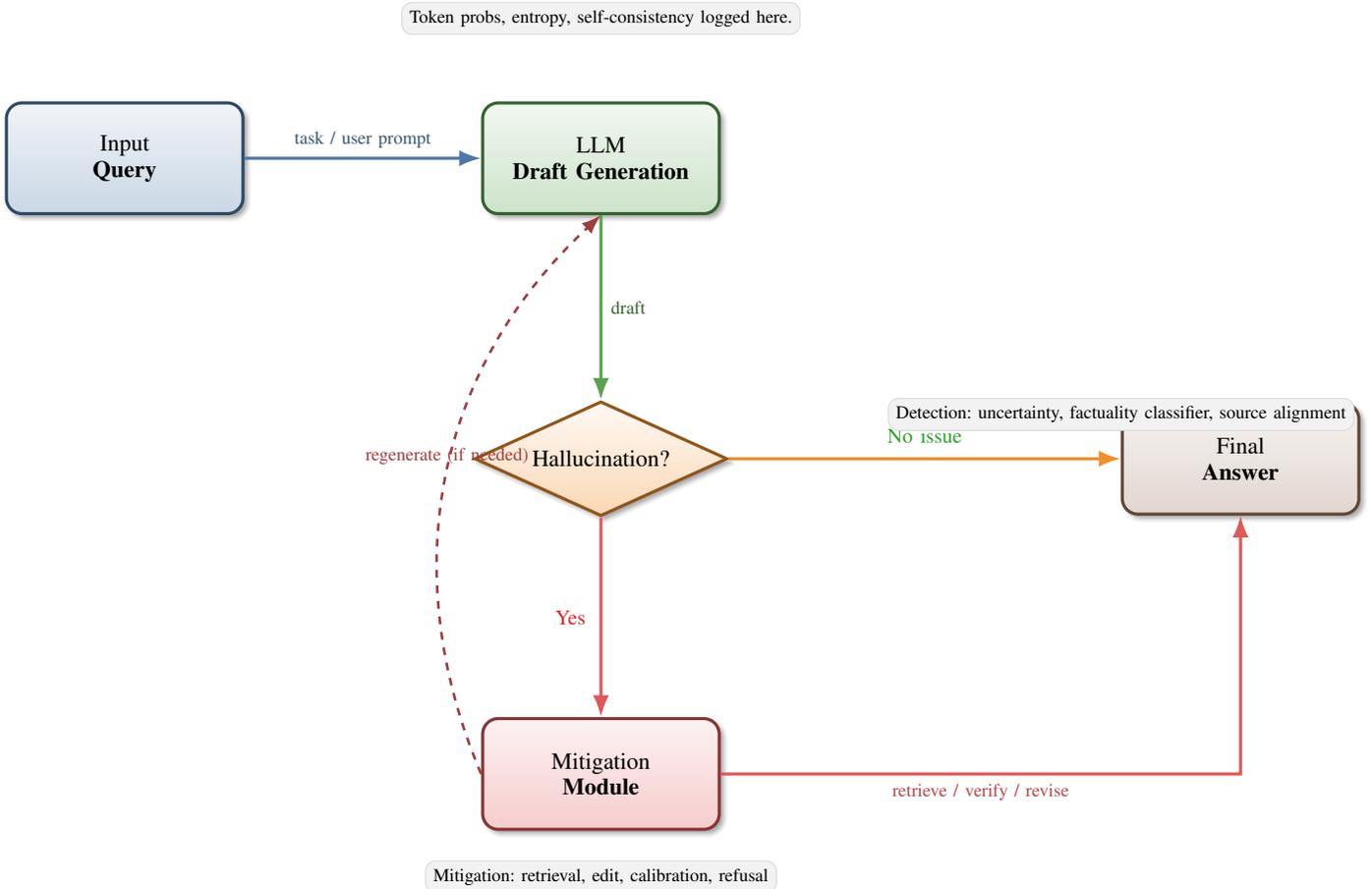
\begin{figure*}[ht]
\centering
\begin{tikzpicture}[node distance=4.2cm and 3.2cm]
  \node[io]            (input)   {Input\\ \textbf{Query}};
  \node[gen, right=of input] (llm)     {LLM \\ \textbf{Draft Generation}};
  \node[decision, below=2.5cm of llm] (check) {Hallucination?};
  \node[final, right=5.3cm of check] (final)   {Final \\ \textbf{Answer}};
  \node[mit, below=2.7cm of check]   (mitigate){Mitigation \\ \textbf{Module}};
  
  \draw[flow=colInput] (input) -- node[above, font=\scriptsize, text=colInput!70!black]{task / user prompt} (llm);
  \draw[flow=colGen]   (llm) -- node[right, font=\scriptsize, text=colGen!60!black]{draft} (check);
  \draw[flow=colDecision] (check) -- node[above, text=colYes, font=\footnotesize, yshift=2pt]{No issue} (final);
  \draw[flow=colMitigate] (check) -- node[left, text=colNo, font=\footnotesize, xshift=-2pt]{Yes} (mitigate);
  \draw[flow=colMitigate] (mitigate) -| node[pos=0.25, below, font=\scriptsize, text=colMitigate!80!black]{retrieve / verify / revise} (final);
  
  \draw[dashedflow=colMitigate!70!black] (mitigate.west) to[bend left=35] node[above, font=\scriptsize, text=colMitigate!70!black]{regenerate (if needed)} (llm.south);
  
  \node[labelbox, above=0.9cm of llm, xshift=-0.0cm] {Token probs, entropy, self-consistency logged here.};
  \node[labelbox, right=0.1cm of check, xshift=2.05cm, yshift=0.6cm] {Detection: uncertainty, factuality classifier, source alignment};
  \node[labelbox, below=0.4cm of mitigate] {Mitigation: retrieval, edit, calibration, refusal};
  
\end{tikzpicture}
\caption{Proposed workflow for hallucination detection and mitigation. The LLM generates a draft response given an input query. A detection module then evaluates the response for potential hallucinations (intrinsic or extrinsic). If no hallucination is detected (No), the response is finalized and returned. If a hallucination is detected (Yes), a mitigation module is invoked, which could involve retrieving additional information, applying corrections, or re-generating portions of the answer. The mitigated answer is then produced as the final answer.}
\label{fig:workflow}
\end{figure*}

Synthesizing the above detection and mitigation strategies, we propose a workflow for deploying an LLM in a setting where factual accuracy is paramount. The workflow ensures that for each user query or input, the system checks for possible hallucinations and applies mitigation steps before finalizing a response. Figure~\ref{fig:workflow} illustrates this pipeline.

The steps in the workflow are:
\begin{enumerate}
    \item \textbf{Initial Generation:} Given a user query or task input, the LLM produces an initial answer. This is done with a potentially cautious decoding strategy (e.g., moderately low temperature to avoid too much randomness).
    \item \textbf{Hallucination Detection:} The draft answer is passed to a detection module. This could involve:
    \begin{itemize}
        \item Checking the model's self-reported uncertainty or the entropy of the generation (from logs recorded during generation).
        \item Using a classifier or heuristic to identify unsupported factual claims (for instance, scanning the answer for sentences containing facts and verifying each against the input or a knowledge source).
        \item If the input provides grounding (like documents), using an overlap or attention-based metric to see if all facts in the answer come from the input.
    \end{itemize}
    If the detection module finds no clear hallucination (the answer appears factual and supported), the pipeline proceeds to output the answer. If a potential hallucination is flagged, we move to the next step.
    \item \textbf{Mitigation Actions:} Upon detecting a likely hallucination, the system can take one or more mitigation actions:
    \begin{itemize}
        \item \textit{Retrieve and Refine:} Perform a retrieval query for the contentious parts of the answer or the whole question. For example, if the answer stated a specific fact that is unverified, query a search engine or database with that fact or question. Incorporate the retrieved evidence into the context and prompt the LLM to regenerate or adjust the answer using the new information (essentially a RAG second-pass).
        \item \textit{Verify and Edit:} Use a fact-checker module to pinpoint which part of the answer is false. Then either programmatically edit the answer (e.g., remove or replace the false statement) or prompt the LLM with feedback. For instance, "In your answer, the statement X seems incorrect. Please correct it." This utilizes the model's ability to do targeted correction when guided.
        \item \textit{Abstain or Qualify:} If the hallucination is due to a question that the model genuinely cannot answer correctly (no knowledge available), the mitigation might be to replace the answer with a refusal or a statement of uncertainty (like "I'm sorry, I don't have that information."). It's better to have no answer than a wrong one in many applications.
    \end{itemize}
    After mitigation, an updated answer is produced.
    \item \textbf{Final Answer:} The refined answer (post-mitigation) is delivered as the final output. Ideally, this answer has any hallucinated content removed or corrected. In cases where retrieval was used, the answer might now explicitly include references or evidence ("According to [source], ...") to increase user trust.
\end{enumerate}

This workflow can be iterative. If the final answer is still uncertain, the detection step could run again. In practice, one iteration is usually aimed for, since too many loops could cause delays. But a system might allow a second loop if the first mitigation still produced an answer that the detector is not fully satisfied with.

An example scenario: User asks a complex question, LLM gives an answer that includes a date and name that detector flags as potentially wrong. The system retrieves relevant Wikipedia info, finds the correct date and name, prompts LLM to correct those. The final answer with correct facts is returned. If the retrieval found no evidence, the system might respond with uncertainty rather than risk a guess.

The above workflow is in line with what some deployed AI assistants do, combining large LLMs with search engines and checkers to reduce incorrect outputs. It leverages both the generative strength of LLMs and the precision of knowledge-based systems.

\section{Evaluation Protocols for Hallucination}
\label{sec:evaluation}
Evaluating hallucination in LLMs requires careful consideration, as it involves assessing factual correctness and faithfulness, which can be subtle. We outline recommended practices for empirically evaluating hallucination and the effectiveness of mitigation techniques.

\subsection{Datasets and Benchmarks}
A variety of benchmarks have been proposed to stress-test models' tendency to hallucinate:
\begin{itemize}
    \item \textbf{Factual Question Answering Benchmarks:} Datasets like TruthfulQA \cite{Lin2021} specifically target whether models produce truthful answers to questions that might prompt common misconceptions or require knowledge. TruthfulQA provides questions and categorizes answers as truthful or false, enabling a direct measure of hallucination (falsehood) rates. Another is the "Factoid QA" where every question has a verifiable answer in Wikipedia (e.g., NaturalQuestions, WebQuestions) - we can check if the model's answer matches the known truth.
    \item \textbf{Summarization Benchmarks with Faithfulness Annotations:} For instance, XSum and CNN/DM summaries have known issues with hallucination. Maynez et al. (2020) annotated XSum model outputs for intrinsic and extrinsic hallucinations \cite{Maynez2020}. Recent datasets (e.g., from Pagnoni et al. 2021 \cite{Pagnoni2021}) include human judgments on factual consistency for many summaries. Using these, one can measure how often a model's summary has unfaithful content and see if mitigation (like a grounded summarizer) lowers that frequency.
    \item \textbf{Knowledge-Grounded Dialogue Benchmarks:} Datasets like Wizard-of-Wikipedia and Holl-E provide dialogues where a model must stick to given knowledge. They often come with metrics like knowledge F1 \cite{Shuster2021} that measure overlap between the model's response and the gold knowledge. A low precision in this overlap means the model introduced content that was not in the knowledge (hallucination).
    \item \textbf{Machine Translation Hallucination Sets:} There are known cases of hallucination in NMT, often with low-resource language pairs. Some research has test sets where source sentences were perturbed or out-of-domain and they check if the translation outputs irrelevant text \cite{Raunak2021}. These can be used to evaluate how often a model produces content not present in the source (an extrinsic hallucination in MT context).
    \item \textbf{Domain-specific factuality tests:} e.g., for medical LLMs, one can use questions from medical exams or factual checks where the answers are known. For coding assistants, hallucination might mean producing code that doesn't compile; there, test suites can catch functional hallucinations.
\end{itemize}

When evaluating a mitigation like RAG or fine-tuning, it's important to test on queries that are challenging and likely to induce hallucination. This can include deliberately out-of-scope questions, ambiguous prompts, or those requiring up-to-date knowledge (which base models might not have learned).

\subsection{Metrics for Hallucination}
We have touched on some metrics earlier, but summarizing:
\begin{itemize}
    \item \textbf{Hallucination Rate / Factuality Score:} The simplest metric is the percentage of outputs that contain a hallucination. This typically requires human evaluation or a highly trusted automatic method. For a given test set, you could count how many answers are fully correct vs have any incorrect info.
    \item \textbf{Intrinsic/Extrinsic Breakdown:} If possible, it is insightful to report the breakdown: e.g., "20\% of summaries had hallucinations: 5\% intrinsic, 15\% extrinsic." This requires labeling each hallucinated case as one or the other, which is usually manual.
    \item \textbf{Knowledge F1 / Content Precision and Recall:} These metrics compare the set of facts in the output to the set in the source or reference. For example, Knowledge F1 \cite{Shuster2021} measures the overlap of factual content (often entities) between a dialogue response and the provided knowledge. A low precision in this overlap means the model introduced content that was not in the knowledge (hallucination).
    \item \textbf{Entailment-based Metrics:} Use a natural language inference (NLI) model to judge if the model's output is entailed by the source (for tasks with source). If the NLI model says the output is not entailed (or contradicted), that's a signal of unfaithfulness. Metrics like FactCC \cite{Kryscinski2020} and other BERT-based classifiers effectively do this; some works fine-tune NLI models specifically for factual consistency.
    \item \textbf{Question-Answering based Metrics (QAGS):} Generate questions from the model’s output, then see if a QA system can answer them correctly using the source text \cite{Wang2020}. If the answers from the source don't match the output, the output likely had unsupported info. This is an indirect but often effective automatic metric for summarization factuality.
    \item \textbf{Human Evaluation Scales:} When possible, use human evaluators to rate outputs on a scale (e.g., 1 to 5) for factual correctness. Define criteria clearly: 5 = no hallucination, fully faithful; 4 = maybe a trivial extrinsic detail added; 3 = some minor incorrect info; 2 = major incorrect info; 1 = almost entirely hallucinated or unusable. This helps gauge severity, not just binary presence.
    \item \textbf{Calibration metrics:} If one aim is to have the model know when it's guessing, one can measure calibration. E.g., the Brier score or Expected Calibration Error (ECE) for the model's predicted probabilities vs actual correctness \cite{Guo2017}. For generative models, a variant might be needed (like measuring if when the model says "I am 90\% sure," it's correct 90\% of the time). Good calibration means fewer unwarranted confident hallucinations.
\end{itemize}

It is often beneficial to use multiple metrics to get a full picture. Automatic metrics can be noisy or one-dimensional, so confirm with some human assessment on a subset.

\subsection{Experimental Settings and Reporting}
We recommend the following when designing experiments to evaluate hallucination:
\begin{itemize}
    \item \textbf{Compare Base vs Mitigated Models:} Always evaluate the original model (without the mitigation strategy) versus the model with the strategy. For example, compare GPT-3 vs GPT-3 + retrieval on the same questions. This directly shows the reduction in hallucination (if any) and helps quantify the benefit.
    \item \textbf{Diverse Test Cases:} Use a mix of easy and hard queries. Some queries that are straightforward factual (which the model likely knows) to establish a baseline of performance, and some deliberately tricky ones. For hallucination study, bias toward those that are challenging.
    \item \textbf{Ablation Studies:} If you introduce a pipeline with multiple components (say retrieval + verification + fine-tuning), perform ablations to see which contribute most. For instance, test retrieval alone, fine-tuning alone, and combined, to evaluate their individual and combined effect on hallucination rate.
    \item \textbf{Measure Impact on Fluency/Other Metrics:} Ensure that efforts to reduce hallucination don’t overly degrade language quality or other desired traits. So, measure something like BLEU/ROUGE for summarization (if applicable) or user satisfaction if possible. In many cases, factuality improvements come with minimal quality loss, but it's good to verify. If a mitigation harms the model's ability to answer at all (maybe it refuses too often), note that trade-off.
    \item \textbf{Statistical significance:} Given the variability in generation, use sufficiently large sample sizes and statistical tests if claiming one method is better. Also consider running multiple trials if using stochastic generation to account for randomness.
    \item \textbf{Error Analysis:} Present a brief analysis of common failure modes even after mitigation. For example, maybe with retrieval the model rarely makes up proper nouns now, but still occasionally mis-states numerical values. Understanding what hallucinations remain can guide future improvements.
\end{itemize}

As an example, an evaluation for a QA model might look like: 500 questions from TruthfulQA, measure truthful answer percent; 100 questions beyond training knowledge (requiring current events info) to see how it handles unknowns; measure ECE of its self-reported confidence; etc., and then compare those metrics before and after applying retrieval + calibration.

In summary, evaluation should be comprehensive, covering both whether hallucinations are reduced and whether the model remains useful and fluent. By following these protocols, researchers can reliably track progress on making LLMs more factual and identify areas that need more work.

\section{Conclusion and Future Work}
\label{sec:conclusion}
Hallucination in large language models remains a significant barrier to their deployment in many real-world scenarios that require reliability and factual accuracy. In this paper, we provided a thorough examination of the problem from theoretical foundations to practical solutions. We formalized what it means for an LLM to hallucinate, distinguishing intrinsic contradictions from extrinsic fabrications, and introduced the concept of hallucination risk as a measurable quantity. We discussed how classical learning theory can bound this risk, yet also noted theoretical results implying that some level of hallucination may be innate for general-purpose models, reinforcing the need for ongoing mitigation efforts \cite{Xu2024}.

On the practical side, we surveyed a spectrum of detection methods (uncertainty-based, calibration-based, and attention-based) and mitigation strategies (RAG, fine-tuning, calibrated decoding, and verification modules). Each approach contributes a piece to the puzzle: retrieval brings grounded knowledge, fine-tuning aligns model behavior with truthfulness, calibration and uncertainty quantification help the model judge when it might be wrong, and verification acts as a safety net to catch mistakes. By integrating these, as illustrated in our proposed workflow, one can build systems that significantly reduce hallucination rates compared to naive LLM usage.

Our recommendations for evaluation serve as a guide to measure progress. It's crucial that the community converges on robust benchmarks and shares best practices for testing factuality. Only through rigorous evaluation can we confidently deploy LLMs in sensitive domains like healthcare, law, or education, where a hallucinated statement could have serious repercussions.

Looking forward, there are several exciting directions for future research. One is \textbf{knowledge boundary estimation}: enabling models to explicitly know and indicate the limits of their knowledge (essentially learning a model of their own ignorance). This could involve the model internally predicting whether it has seen sufficient evidence for a query or if it should defer to an external source \cite{Ji2023}. Another direction is \textbf{dynamic retrieval and reasoning}, where models not only fetch facts but also perform reasoning steps (e.g., using chain-of-thought prompting combined with tool use) to ensure consistency and correctness of multi-hop answers. Advances in \textbf{multi-modal grounding} may also help; for instance, linking text to images or databases to cross-verify information could reduce hallucination (like verifying a generated caption against the actual image content).

From a theoretical standpoint, developing more refined frameworks to analyze why and when hallucinations occur could inform training regimes. Could we characterize certain training distributions or model architectures that inherently minimize hallucinations The intersection of causal inference and LLM training might offer insights into how models pick up spurious facts and how to mitigate that.

In conclusion, while hallucination in LLMs is a challenging problem, the combination of theoretical understanding and a multifaceted engineering approach provides a promising path to taming it. By continuing to ground models in reality, encourage them to know what they don't know, and rigorously checking their outputs, we move closer to LLMs that can be both creative and consistently truthful. Such models will greatly enhance trust and broaden the safe applicability of AI in society.

\bibliographystyle{IEEEtran}

\end{document}